\documentclass[10pt,twocolumn]{article}

\PassOptionsToPackage{table,xcdraw}{xcolor}

\usepackage[utf8]{inputenc}
\usepackage[english]{babel}
\usepackage{amsmath,amssymb,amsthm}
\usepackage{float}
\usepackage{caption}
\usepackage{subcaption}
\usepackage{booktabs}
\usepackage{siunitx}

\usepackage{xcolor}
\usepackage{authblk}
\usepackage{geometry}
\usepackage{fancyhdr}
\usepackage{titlesec}
\usepackage{courier}
\usepackage{tcolorbox}
\usepackage{dblfloatfix} 
\usepackage{placeins} 

\usepackage{algorithm}
\usepackage{algorithmic}

\usepackage{graphicx}
\usepackage{multirow}
\usepackage{hyperref}
\usepackage{makecell}

\usepackage{natbib}
\setcitestyle{authoryear,open={(},close={)}}

\renewcommand\thesection{\Roman{section}}
\renewcommand\thesubsection{\Alph{subsection}}
\titleformat{\section}[block]{\large\scshape\centering}{\thesection.}{1em}{}
\titleformat{\subsection}[block]{\large}{\thesubsection.}{1em}{}

\title{\fontsize{16pt}{10pt}\selectfont\textbf{Configuration Interaction Guided Sampling with Interpretable Restricted Boltzmann Machine}}

\author[1,2]{Jorge I. Hernandez-Martinez$^{\dagger}$}
\author[1]{Andres Mendez-Vazquez}
\author[2]{Gerardo Rodriguez-Hernandez$^{\dagger}$}
\author[1,2]{Sandra Leticia Ju\'arez-Osorio}

\affil[1]{\small CINVESTAV Guadalajara, Department of Electrical Engineering and Computer Science, Jalisco, 45017, Mexico}
\affil[2]{\small Tecnologico de Monterrey, School of Engineering and Sciences, Jalisco, 45138, Mexico}
\affil[ ]{\small $^{\dagger}$Corresponding Authors: \href{mailto:jivan.hernandez@cinvestav.mx}{jivan.hernandez@cinvestav.mx}, \href{mailto:gerardo.rodriguez@tec.mx}{gerardo.rodriguez@tec.mx}}

\date{}

\begin{document}
	\maketitle
	\thispagestyle{fancy}
	
	\begin{abstract}
		\noindent
		We propose a data-driven approach using a Restricted Boltzmann Machine (RBM) to solve the Schrödinger equation in configuration space. Traditional Configuration Interaction (CI) methods construct the wavefunction as a linear combination of Slater determinants, but this becomes computationally expensive due to the factorial growth in the number of configurations. Our approach extends the use of a generative model such as the RBM by incorporating a taboo list strategy to enhance efficiency and convergence. The RBM is used to efficiently identify and sample the most significant determinants, thus accelerating convergence and substantially reducing computational cost. This method achieves up to 99.99\% of the correlation energy while using up to four orders of magnitude fewer determinants compared to full CI calculations and up to two orders of magnitude fewer than previous state of the art methods.  Beyond efficiency, our analysis reveals that the RBM learns electron distributions over molecular orbitals by capturing quantum patterns that resemble Radial Distribution Functions (RDFs) linked to molecular bonding. This suggests that the learned pattern is interpretable, highlighting the potential of machine learning for explainable quantum chemistry
		
		\vspace{1mm}
	\end{abstract}
	
	\section{Introduction}
	\label{sec:Introduction}
		Understanding the quantum properties of molecules and materials is essential across many scientific and technology fields. While the Schrödinger equation forms the theoretical foundation for this understanding, its exact solution is often intractable for more than a few electrons. Configuration Interaction (CI) \citep{KARAZIJA2013131,Bunge_2010,ci_article_adnan} methods offer a powerful approximation by expanding the wavefunction in a basis of electronic configurations. However, the number of determinants grows factorially with the number of electrons and molecular orbitals, which limits their practical applicability.
		
		To address this challenge, we propose a novel approach that leverages Restricted Boltzmann Machines (RBMs) to efficiently sample the configuration space, focusing on those determinants with the highest contribution to the solution. RBMs are probabilistic neural networks with a simple two-layer structure that can learn complex, nonlinear distributions. Unlike other generative models such as Generative Adversarial Networks (GANs) or Variational Autoencoders (VAEs), the RBM's simple structure allows for efficient sampling and makes it easier to interpret the learned features. Our model extends the use of the RBM for wavefunction solutions \citep{articulo_rbm_carleo, basile_herzog}, building upon their generative capabilities to enhance sampling efficiency and interpretability.
		
		The key innovation of our approach lies in its iterative, guided model training strategy. At each iteration, determinants with higher contributions (as governed by Slater's rules) are prioritized and retained, while a taboo list prevents the re-selection of low impact determinants. This not only reduces the number of determinants required by up to two orders of magnitude but also keeps the computational cost manageable. Additionally, the method provides interpretability: The RBM appears to uncover distinctive patterns in the distribution of determinants that correlate with physical properties such as orbital energies and radial probability distributions.
		
		In summary, our study introduces a significant advance in computational quantum chemistry by integrating a guided RBM-based CI strategy with a practical implementation, achieving a reduction in computational cost compared to traditional and state of art CI methods. The remainder of the paper is organized as follows. Section 2 provides the theoretical background on solving the Schrödinger equation and reviews the fundamentals and limitations of CI methods, along with an overview of RBMs. In Subsection \ref{Proposed_Approach}, we detail our proposed model and its key innovations. Section \ref{Results_and_Discusion} presents the results and discussion, and Section \ref{Conclusions} concludes the paper.

		\section{Methodology}
		\label{sec:Material and methods}
		
		\subsection{Theoretical Background}
		\label{sec:Theoretical_Background}
		
		In quantum mechanics, the time-independent Schrödinger equation,
		\[
		H\psi = E\psi,
		\]
		provides the fundamental framework to predict a system’s properties. Here, \(H\) is the Hamiltonian operator, \(\psi\) is the many-electron wave function, and \(E\) represents the energy eigenvalue. For many-body systems, an accurate Hamiltonian must account for the kinetic energies of electrons and nuclei as well as their mutual interactions. Under the Born–Oppenheimer approximation, the nuclei are treated as fixed, leading to a Hamiltonian of the form:
		\[
		\left[\frac{\hbar^2}{2m}\sum_{i=1}^{N} \nabla_i^2 + \sum_{i=1}^{N} V(r_i) + \sum_{i=1}^{N} \sum_{j<i} U(r_i,r_j)\right]\psi = E\psi.
		\]
		In this equation, the first term represents the kinetic energy of each electron, the second term \(V(r_i)\) describes the Coulomb attraction between electrons and nuclei,
		\[
		\sum_{i=1}^{N} V(r_i) = \sum_{i=1}^{N} \sum_{j=1}^{N_{\text{atoms}}} \frac{-Z_j}{|r_i-R_j|},
		\]
		and the third term \(U(r_i,r_j)\) accounts for electron-electron repulsion:
		\[
		\sum_{i=1}^{N} \sum_{j<i} U(r_i,r_j)= \sum_{i=1}^{N} \sum_{j<i} \frac{e^2}{|r_i-r_j|}.
		\]
		
		Due to the rapidly increasing dimensionality of the wave function, the calculation of an exact solution of the Schrödinger equation becomes computationally intractable. For instance, a CO\(_2\) molecule involves a 66-dimensional wave function, while a nanocluster of 100 Pt atoms may require over 23,000 dimensions \citep{libro_density_practical}.
		
		To overcome this “curse of dimensionality,” several approximation methods have been developed. For example, one common strategy is to use wave function-based approaches that improve upon a reference state. This reference is typically provided by the Hartree–Fock (HF) method, which approximates the many electron problem by representing the system with a single Slater determinant and treating electron–electron interactions through a mean field potential. However, HF neglects dynamic electron correlation and often provides only a qualitative description of the system \citep{marti_2025,george_2009,tubman_2016}.
		
		To account for these correlation effects, post Hartree–Fock methods expand the wave function as a linear combination of multiple electronic configurations, also known as configuration state functions (CSFs):
		\begin{equation}
			\Psi = \sum_{i=0}^{I} c_i \Phi_i = c_0 \Phi_0 + c_1 \Phi_1 + \ldots,
			\label{equation_ci}
		\end{equation}
		where each CSF \(\Phi_i\) is weighted by a coefficient \(c_i\) that indicates its contribution to the total wave function. Importantly, these coefficients are not uniformly distributed; only a small subset of determinants contribute in a significant way. Thus, identifying these key determinants is essential for achieving a good approximation with fewer configurations.

		Truncated Configuration Interaction (CI) methods form a hierarchy of approximations based on limiting the excitation rank relative to the Hartree–Fock reference. For example, CISD includes only single and double excitations, while CISDTQ extends this to triple and quadruple excitations. Each successive level captures increasingly subtle electron correlation effects. For example, CISD typically recovers about 95\% of the correlation energy in small molecules at equilibrium geometries \citep{cisd_corelation_energy,BLINDER20191}, while CISDTQ can approach 99.8\% of the full CI energy \citep{cisdtq_corelation_energy}. At the top of this hierarchy lies Full Configuration Interaction (FCI), which includes all possible excitations within a given one-particle basis and provides, in principle, an exact solution to the Schrödinger equation in that space. However, the factorial growth in the number of determinants with system size, makes FCI computationally infeasible for anything beyond the smallest molecules.
		
		Coupled Cluster methods, particularly coupled cluster singles doubles perturbative triples CCSD(T), also rely on a determinant based expansion, but use an exponential ansatz that offers size extensivity and excellent accuracy for ground state properties. CCSD(T) is often regarded as the "gold standard" in quantum chemistry due to its reliable performance across a wide range of molecular systems \citep{bartlett2007coupled}. However, it remains computationally expensive, with a scaling of $\mathcal{O}(N^7)$ with respect to system size, and it may fail in situations involving strong static correlation or near degenerate electronic states, such as bond dissociation or transition metal complexes \citep{couple_cluster_2019, couple_cluster_2015}

		Another notable approach is the Configuration Interaction using a Perturbative Selection made Iteratively (CIPSI). CIPSI iteratively adds determinants to the Hamiltonian by selecting those with the largest second order perturbative contributions, thus efficiently capturing the most important correlation effects without exhaustively exploring the entire configuration space. However, CIPSI relies heavily on perturbation theory approximations, which can lead to inaccuracies when describing strongly correlated systems or when the second-order perturbative corrections become large, thus limiting its reliability and computational efficiency in some challenging cases \cite{dash_2021}.

		Alternatively, density-based methods such as Density Functional Theory (DFT) \citep{cohen_dft_limitations} offer computational efficiency but may suffer from systematic errors (e.g., underestimating reaction barriers or band gaps \citep{szabo1996modern}).

		Because the most important contributions to \(\Psi_{\text{FCI}}\) are concentrated in a small subset of determinants, it is crucial to develop methods that can selectively sample and include only those determinants. This motivates our approach, which uses artificial intelligence, specifically, a RBM to guide the selection of determinants, thereby reducing computational cost while preserving accuracy. In the following subsection, we provide a brief overview of the theoretical foundations of FCI and the RBM model that supports our method.

		\subsection{Background on FCI and RBM}
		\label{sec:Background_FCI_RBM}

		In electronic structure calculations, solving the Schrödinger equation is often done using well established methods that represent the many electron wavefunction as a linear combination of electronic configurations. While the Hartree Fock method simplifies the problem by assuming that each electron moves independently in a central potential,  neglecting the explicit \(r_i\) term in the many body Hamiltonian and limiting accuracy due to the use of an average potential for electron-electron interactions \citep{BLINDER20191}. Therefore CI methods overcome this limitation by explicitly accounting for electron correlation.Thus, CI constructs properly antisymmetric wave functions by combining multiple electronic configurations, typically represented by excited Slater determinants (Equation \ref{equation_ci}). Each Slater determinant is an antisymmetric function of spin orbitals, ensuring compliance with the Pauli exclusion principle \citep{libro_teoria_fci_david_shrerrill, libro_modern_quantum_chemistry}. In practice, the Hamiltonian matrix is constructed by evaluating its elements using Slater’s rules, which require the computation of one and two-electron integrals:
		\begin{equation}
			\langle i | \hat{h} | j \rangle = \int \phi_i^*(\mathbf{r}_1) \hat{h}(\mathbf{r}_1) \phi_j(\mathbf{r}_1) \, d\mathbf{r}_1,
		\end{equation}
		\begin{equation}
			\langle ij | kl \rangle = \int \phi_i^*(\mathbf{r}_1) \phi_j^*(\mathbf{r}_2) \frac{1}{r_{12}} \phi_k(\mathbf{r}_1) \phi_l(\mathbf{r}_2) \, d\mathbf{r}_1\,d\mathbf{r}_2,
		\end{equation}
		which populate the Hamiltonian elements 
		\[
		H_{ij} = \langle \Phi_i | \hat{H} | \Phi_j \rangle,
		\]
		that are then diagonalized to obtain the coefficients \(c_i\) (see Eq. \ref{equation_ci}) \citep{libro_teoria_fci_david_shrerrill, LESTRANGE2018295}.
		
		Although the factorial growth of the determinant space is well known, only a small subset of configurations contributes significantly to the wavefunction, as the coefficients $c_i$ are highly not uniformly distributed. Each configuration represents a unique arrangement of electrons among the available molecular orbitals, whose number depends on the chosen basis set. For instance, in a closed shell system, the configurations correspond to the possible distributions of alpha (spin up) and beta (spin down) electrons, always obeying the Pauli exclusion principle (i.e., no two electrons with the same spin can occupy the same orbital). As a result, the full configuration space quickly becomes intractable. For example, the estimated number of determinants is given by
		\[
		N_{det} \approx \binom{N_o}{N_e/2}^2,
		\]
		where \(N_o\) is the number of molecular orbitals and \(N_e\) is the total number of electrons. This estimation leads to the values shown in Table \ref{tabla_num_dets_fci} for simple molecules.
		
		\begin{table}[H]
			\centering
			\begin{tabular}{|c|c|c|c|}
				\hline
				System & Basis & Active Space                                & FCI $N_{dets}$ \\ \hline
				\rowcolor[HTML]{EFEFEF} 
				$C_2$   & 6-31G                      & (12 elecs, 18 orbs)                         & \num{3.35e8}                               \\
				\rowcolor[HTML]{FFFFFF} 
				$N_2$   & 6-31G                      & (14 elecs, 18 orbs)                         & \num{1.01e9}                               \\
				\rowcolor[HTML]{EFEFEF} 
				$H_2O$  & 6-31G                      & \cellcolor[HTML]{EFEFEF}(10 elecs, 13 orbs) & \num{1.66e6}                               \\
				\rowcolor[HTML]{FFFFFF} 
				$C_2$   & cc-pVDZ                    & \cellcolor[HTML]{FFFFFF}(8 elecs, 26 orbs)  & \num{2.24e8}      \\
				\rowcolor[HTML]{EFEFEF} 
				$N_2$   & cc-pVDZ                    & \cellcolor[HTML]{EFEFEF}(10 elecs, 26 orbs) & \num{4.33e9}       \\
				\rowcolor[HTML]{FFFFFF} 
				$H_2O$  & cc-pVDZ                    & \cellcolor[HTML]{FFFFFF}(8 elecs, 23 orbs)  & \num{7.84e7}      
			\end{tabular}
			\caption{Required number of determinants for each molecule using different basis sets in FCI calculations}
			\label{tabla_num_dets_fci}
		\end{table}

\begin{figure}[h]
			\begin{center}
				\includegraphics[width=.15\textwidth]{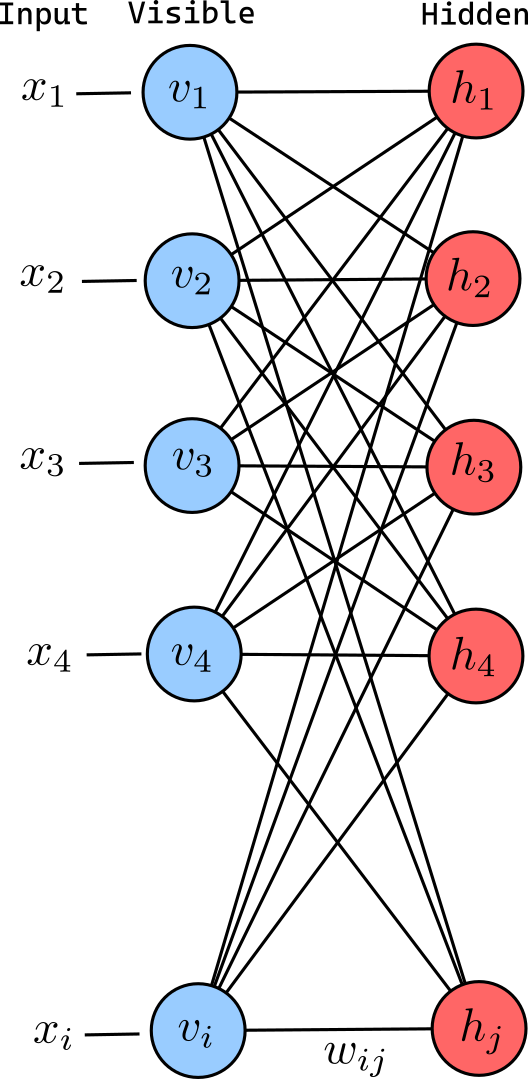}
				\caption{General architecture of the Restricted Boltzmann Machine.}
				\label{fig_rbm}
			\end{center}
		\end{figure}

		To address this issue, our approach leverages an RBM to guide the sampling process. An RBM is an unsupervised generative model that learns complex, non-linear probability distributions over the configuration space. It consists of two layers of nodes: a visible layer that represents the observed data and a hidden layer that captures dependencies between variables (Figure~\ref{fig_rbm}).
		
		The model is defined by its energy function:
		\begin{equation}
			E(v, h) = -\sum_{i} a_i v_i - \sum_{j} b_j h_j - \sum_{i,j} v_i h_j w_{ij},
		\end{equation}
		which determines a joint probability distribution over visible ($v$) and hidden ($h$) units:
		\begin{equation}
			p(v,h)=\frac{1}{Z} e^{-E(v, h)},
			\label{equation_rbm_join_probability}
		\end{equation}
		with $Z$ being the partition function \citep{articulo_overview_rbms, Hinton2012}.
		Here, $a_i$ and $b_j$ are the biases associated with visible and hidden units respectively, and $w_{ij}$ represents the weight connecting visible unit $i$ to hidden unit $j$.
		
		RBMs are trained using Contrastive Divergence \citep{Hinton2012}, which enables efficient learning of the model parameters by approximating the gradient of the log-likelihood. This training procedure helps the RBM to quickly capture the underlying probability distribution of the input data without the need for exhaustive sampling of the partition function.
		
		To sample from the model, the RBM alternates between updating hidden and visible units based on conditional activation probabilities. These are given by:
		\begin{align}
			p(h_j = 1 \mid v) &= \sigma\left(\beta \left(b_j + \sum_i v_i w_{ij}\right)\right), \label{ecuacion_prob_to_hidden} \\
			p(v_i = 1 \mid h) &= \sigma\left(\beta \left(a_i + \sum_j h_j w_{ij}\right)\right), \label{ecuacion_prob_to_visible}
		\end{align}
		where $\sigma(x) = 1/(1 + e^{-x})$ is the logistic sigmoid function. These equations define the Gibbs sampling process used during training, where $\beta$ is the inverse temperature. This parameter controls the level of stochasticity: Low $\beta$ values (high temperatures) lead to more random behavior, while high $\beta$ values result in sharper, more deterministic activations.

		The true strength of the RBM lies in its representational capacity. In theory, an RBM can learn to approximate a wide range of probability distributions. In fact, increasing the number of hidden units improves performance by enhancing the training log likelihood, and reducing the Kullback-Leibler divergence between the data distribution and the model’s distribution \citep{representational_power_rbms}. It has been shown that an RBM with \(k+1\) hidden units can closely approximate any distribution over \(\{0,1\}^n\), where \(k\) denotes the number of input vectors with non zero probability. Moreover, it has been demonstrated that an RBM with \(2^n-1\) hidden units can approximate any distribution \citep{article_refinements_rbms}, and further refinements suggest that RBMs with \(\alpha2^n-1\) hidden units (with \(\alpha < 1\)) are sufficient for a broad class of distributions \citep{article_hierarchical_models}.
		
		In the context of this work, such representational power is crucial: The RBM efficiently captures complex distributions over electronic configurations and enables targeted sampling of the most relevant determinants. By prioritizing configurations with higher contributions to the wave function, the RBM guided approach significantly reduces the computational cost of FCI calculations, compared to traditional Monte Carlo methods. For a more detailed discussion on the functioning and training of RBMs, please refer to Appendix A.

		Previous works have utilized RBMs in quantum many-body problems, but with different objectives. Carleo et al.~\cite{articulo_rbm_carleo} employed RBMs to parameterize the wavefunction itself, approximating $\psi$ through the joint probability distribution of visible and hidden units (Equation \ref{equation_rbm_join_probability}). This approach does not generate new determinants, but instead represents the wavefunction compactly. Basile et al.~\cite{basile_herzog} trained an RBM using a randomly sampled dataset from CISD to assist in determinant selection, incorporating explicit coefficient weights to prioritize determinants and removing highly dominant ones like Hartree-Fock. Both methods offer significant insights, yet face limitations: Carleo's approach does not directly identify new determinants beyond the fixed basis, and Basile's method relies heavily on biased coefficient sampling. Addressing these limitations motivates our proposed strategy, described next.

		\section{Proposed Approach}
		\label{Proposed_Approach}
		
		This section details the implementation of our RBM-guided sampling strategy to optimize the sampling process in the FCI method. Building upon the RBM’s ability to model complex configuration distributions, we introduce an iterative sampling mechanism that progressively refines the determinant set. Unlike traditional methods such as Hartree–Fock or CISD, which rely on a fixed and often truncated determinant space, our method dynamically generates and expands the set of Slater determinants across virtual orbitals. By focusing on configurations with higher predicted contributions, each iteration yields increasingly accurate wavefunction approximations while avoiding redundant exploration.
		
		For consistency in the initial setup and energy evaluation, we employ \textit{Quantum Package}~\cite{article_quantum_package} to:
		\begin{enumerate}
			\item Compute the self-consistent Hartree–Fock wavefunction.
			\item Obtain the CISD determinants, which serve as the initial ansatz for our iterative approach.
			\item Perform diagonalization of the Hamiltonian matrix at each iteration using the generated determinants, employing the Davidson method to efficiently obtain the ground state eigenvalue.
		\end{enumerate}
		
		The full algorithm is summarized in Algorithm~\ref{alg:generative_pruning_summary}. The process begins with an initial set of CISD determinants and iteratively refines the determinant pool through RBM-guided sampling. On each iteration, the RBM is trained to learn an approximate distribution over the configuration space. New determinants are then generated based on this learned distribution and combined with those from previous iterations.

\begin{figure*}[t]
			\centering
			\includegraphics[width=.6\textwidth]{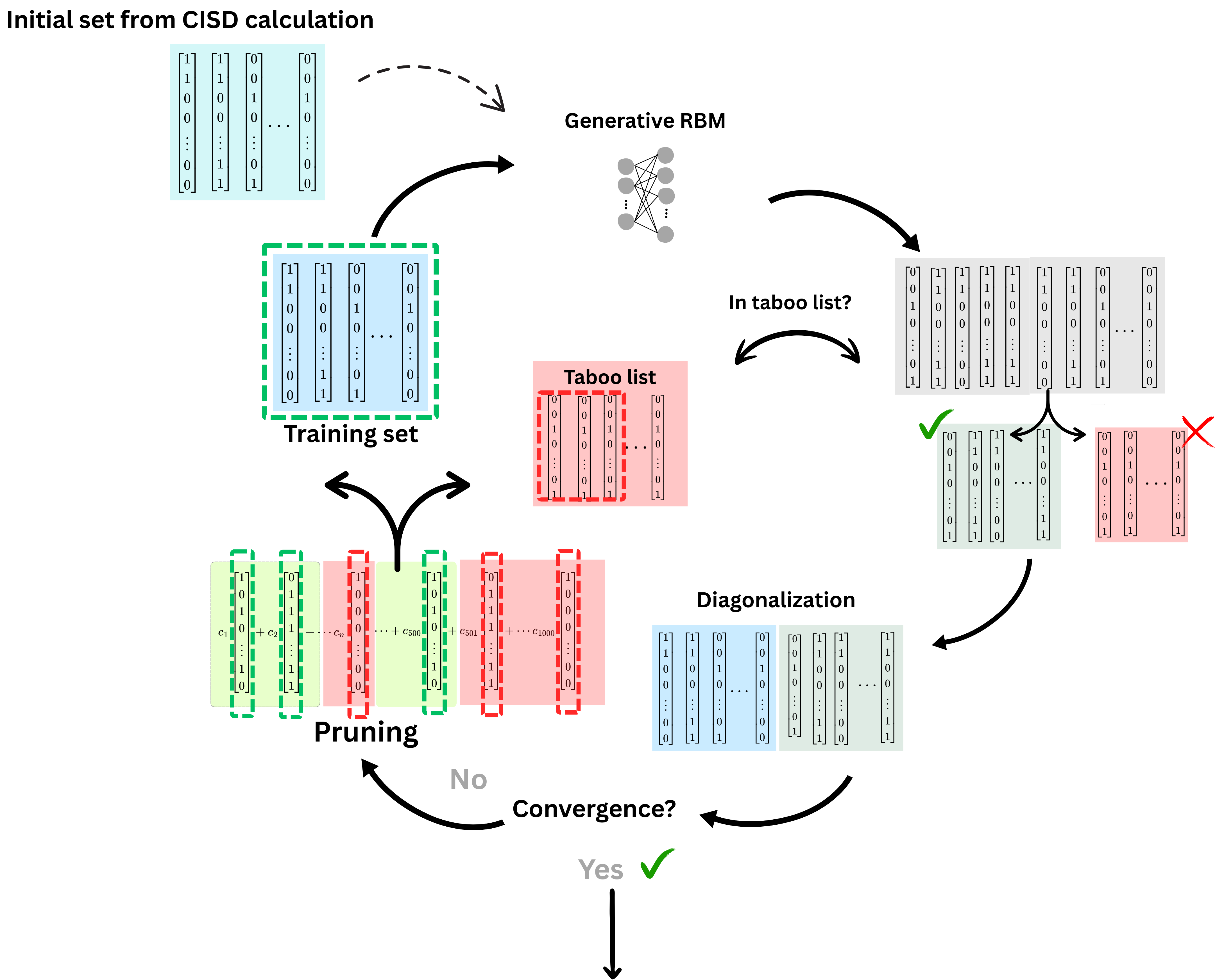}
			\caption{Overview of the RBM guided sampling and pruning cycle, including the use of a taboo list to avoid redundant configurations.
			}
			\label{fig:workflow_diagram}
		\end{figure*}

		After removing duplicates and previously discarded configurations, the Hamiltonian is built and diagonalized to obtain the CI coefficients. Determinants whose coefficients fall below a predefined threshold \texttt{Pthreshold} are pruned and stored in a taboo list to avoid re-selection in future iterations. The pruning of determinants is performed based on their squared absolute coefficient values ($|c_i|^2$), as these reflect the probability weight of each configuration in the wavefunction expansion. This physically motivated choice ensures that only the most relevant configurations, those significantly contributing to the probability distribution, are retained. Consequently, it avoids systematic overestimation of the ground state energy and ensures that the wavefunction remains accurate. Then the determinant pool is updated based on the remaining significant determinants, and the Hamiltonian matrix is re-evaluated with this new set of configurations. This process repeats until the total energy change between consecutive iterations falls below a stability threshold \texttt{Sthreshold}, indicating convergence.
		A high-level overview of the proposed approach is illustrated in Figure \ref{fig:workflow_diagram}.
		
		Each iteration refines the selection of relevant determinants, yielding an increasingly accurate approximation of the wavefunction. Importantly, as the configuration space grows with each iteration, the resulting wavefunction approaches the Full CI solution within the chosen basis. This property stems from the fact that the CI wavefunction spans a subspace of the Hilbert space, and its accuracy systematically improves as more determinants are included~\citep{szabo1996modern, libro_teoria_fci_david_shrerrill,Shepherd_2012}. 
		
		Unlike other approaches that explicitly exclude highly dominant configurations from the training data, such as Hartree–Fock determinants to avoid over representation, our method does not impose any such bias. All determinants, regardless of their coefficient magnitude, are included during training. This allows the RBM to learn an unbiased probability distribution over the configuration space, leading to a more complete and physically representative model of the wavefunction.
		
		As a result, our method achieves high quality approximations while using significantly fewer determinants than traditional approaches.
		
		\begin{algorithm}[H]
			\caption{Generative Model with Determinant Pruning}
			\label{alg:generative_pruning_summary}
			\begin{algorithmic}
				\STATE \textbf{Define constants:}
				\STATE $\text{max\_iterations} \in \mathbb{N}$
				\STATE $\text{Pthreshold} \in \mathbb{R}$ \COMMENT{Coefficient pruning threshold}
				\STATE $\text{Sthreshold} \in \mathbb{R}$ \COMMENT{Energy convergence threshold}
				\STATE $\text{epochs} \in \mathbb{N}$
				
				\STATE \textbf{Initialize variables:}
				\STATE $E \leftarrow 0$ \COMMENT{Initial energy}
				\STATE $\text{PDetsList} \leftarrow \emptyset$ \COMMENT{List of pruned determinants}
				\STATE $\text{dets} \leftarrow \text{Initial CISD calculation}$
				
				\FOR{iteration $ = 1$ to $\text{max\_iterations}$}
				\FOR{epoch $ = 1$ to $\text{epochs}$}
				\STATE Train RBM (dets)
				\ENDFOR
				\STATE $\mathbf{D}'\leftarrow \text{GenerateDeterminants(RBM)}$
				\STATE $\mathbf{D}'\leftarrow \text{$\mathbf{D}'$} + \text{dets}$
				
				\STATE $\mathbf{D}' \leftarrow \text{Unique($\mathbf{D}'$)}$
				\STATE $\mathbf{D}' \leftarrow \text{CheckPrunedDets($\mathbf{D}'$,PDetsList)}$
				\STATE $\mathbf{C} \leftarrow \text{Diagonalization($\mathbf{D}'$)}$ 
				\STATE $ \text{dets} \leftarrow \text{$\mathbf{D}'$[$\mathbf{C} >$ Pthreshold ]}$
				\STATE $\text{PDetsList} \leftarrow \text{PDetsList} \cup \{\mathbf{D}'[i] \mid \mathbf{C}[i] \leq \text{Pthreshold}\}$
				\STATE $E_{\text{new}} \leftarrow \text{ComputeEnergy}(\text{dets})$
				\IF{$|E_{\text{new}} - E| < \text{Sthreshold}$}
				\STATE \textbf{break}
				\ENDIF
				\STATE $E \leftarrow E_{\text{new}}$
				\ENDFOR
			\end{algorithmic}
		\end{algorithm}

		\subsection{Innovations and Implementation Details}
		\label{Innovations_and_Implementation_Details}
		
		Our approach introduces several innovations to enhance the efficiency and relevance of determinant selection in the RBM-guided sampling process. First, we employ a taboo list to prevent the re-selection and re-diagonalization of previously pruned determinants, reducing redundant computations and significantly improving efficiency. Second, while the RBM learns the probability distribution of relevant determinants, some generated determinants may differ from those in the current set by more than two spin orbitals. According to Slater’s rules, the Hamiltonian matrix element for such determinants will be zero, meaning they do not directly connect to the existing determinant set, and therefore do not contribute meaningfully to the ground state energy. Our approach explicitly filters out these unconnected determinants by computing their Hamming distance with respect to the existing determinant set. Ensuring that only determinants with Hamming distances of at most four (allowing up to two differences per alpha and beta spin) are retained. This filtering step ensures a connected determinant space, allowing our model to reach an accurate solution with fewer determinants.
		
		Furthermore, we go beyond determinant selection by analyzing the learned distributions across different molecules and basis sets, revealing patterns that were previously unexamined.
		To further optimize efficiency, we employ a binary-to-decimal conversion strategy~\cite{osti_1633273} for determinant uniqueness verification, enabling fast vectorized operations and streamlined data handling.
		
		Finally, to improve the selection process, we implement a non-uniform finite distribution sampler~\cite{libro_algothms_and_computations} that biases electron permutations based on the RBM learned distribution. This ensures that determinant generation remains physically meaningful while optimizing computational performance.
		
		By integrating these innovations, our method achieves high accuracy with significantly fewer determinants compared to traditional FCI or perturbatively selected CI methods like  CIPSI. The next section evaluates the method’s performance, analyzing both its computational efficiency and the distribution it learns.

		\subsection{Performance and Convergence Analysis}
		\label{Performance_and_Convergence_Analysis}
		Our pruning method based on squared coefficients ($|c_i|^2$) systematically refines determinant selection, progressively enhancing the accuracy of the ground state energy approximation.
		
		Furthermore, although previously pruned determinants are excluded from subsequent iterations via the taboo list, their future relevance cannot be completely dismissed. In principle, new determinants generated in later iterations might establish connections with previously discarded determinants, potentially increasing their importance in the wavefunction expansion. Nevertheless, this scenario is unlikely due to two main reasons: Firstly, we use a very low pruning threshold, ensuring that only configurations with negligible contributions are discarded; secondly, our iterative method naturally progresses towards generating determinants with higher order excitations, which typically have lower contributions (weights) to the overall wavefunction. Thus, any new connections involving previously pruned determinants would most likely involve high energy (low weight) configurations, making significant reconnection improbable. 
		
		As a result, our pruning strategy not only ensures accurate determinant selection at each iteration but also avoids unnecessary recomputation, contributing to overall computational efficiency. Rather than re-evaluating discarded configurations, each iteration focuses on refining the current selection, promoting a more effective exploration of the determinant space. This iterative refinement also keeps the computational cost of diagonalization within manageable limits, maintaining the expected $O(n^2)$ or at worst $O(n^3)$ scaling of the Davidson algorithm during the diagonalization step.

		\section{Results and Discussion}
		\label{Results_and_Discusion}
\begin{figure*}[t]
			\begin{center}
				\includegraphics[width=0.7\textwidth]{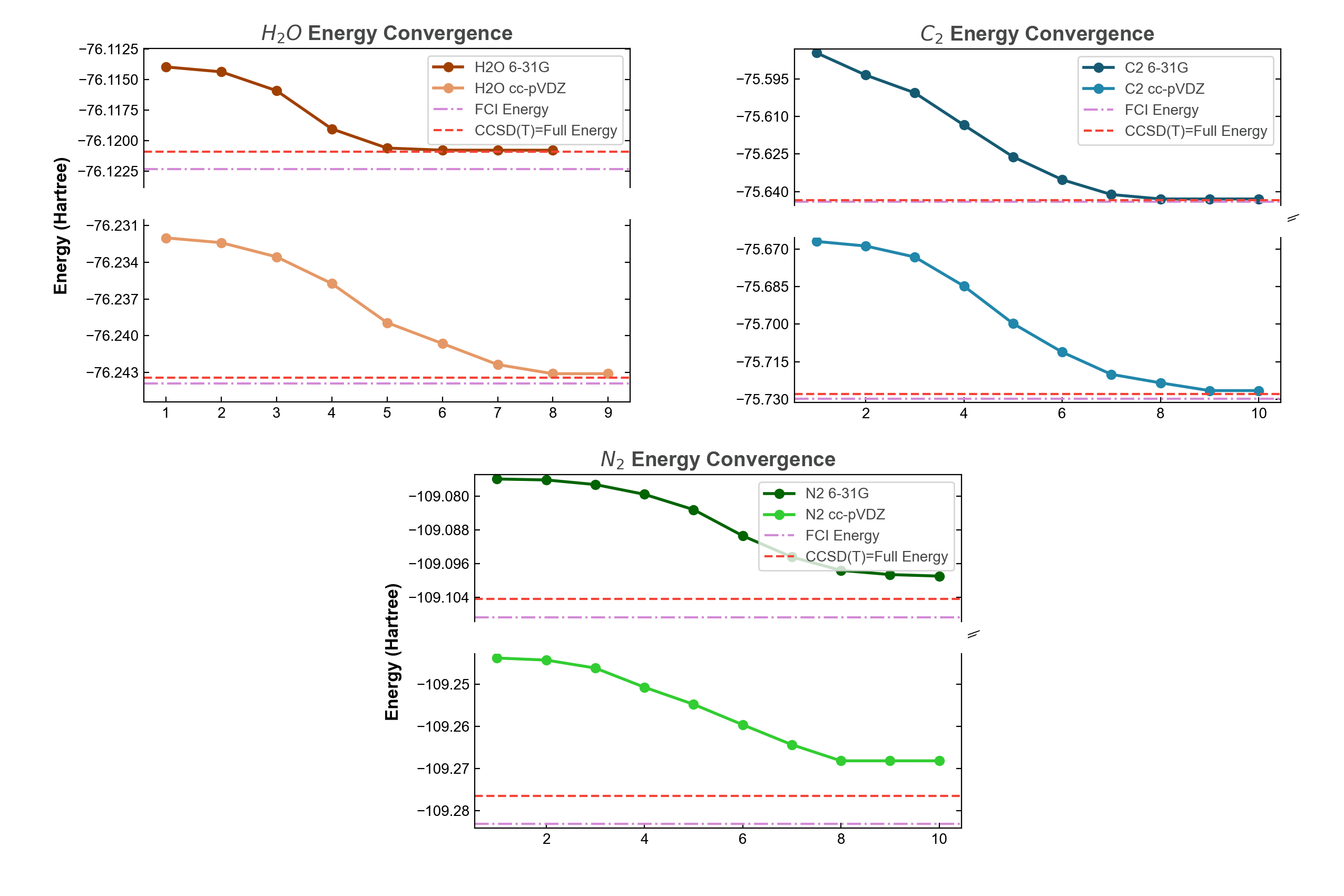}
				\caption{Energy convergence across iterations for molecules of  $H_2O$, $N_2$, and $C_2$ in cc-pVDZ and 631G basis.}
				\label{all_energy_converges}
			\end{center}
		\end{figure*}

		We evaluated our model on $H_2O$, $N_2$, and $C_2$ using the 6-31G and cc-pVDZ basis sets. Figure \ref{all_energy_converges} shows the energy convergence across iterations, revealing that our method consistently approaches FCI energy levels. Notably, the final energies also fall within the range of CCSD(T) benchmarks, confirming that our model delivers chemically accurate results while being computationally more efficient.

		Table \ref{energy_comparison} provides a detailed comparison between CISD, FCI, CCSD(T), and our approach. Across all systems, our method achieves an accuracy of at least 99.99\% of the FCI energy after a small number of iterations, using significantly fewer determinants.

		\begin{table*}[t]
			\centering
			\resizebox{\textwidth}{!}{ 
				\renewcommand{\arraystretch}{1.2} 
				\setlength{\tabcolsep}{10pt} 
				\begin{tabular}{c c c c c c c c c}
					\hline
					\textbf{Molecule} & \textbf{Basis} & \textbf{CISD Energy} & \textbf{FCI Energy} & \textbf{Our Energy} & \textbf{\makecell{CCSD(T)\\ Energy}} & \textbf{\makecell{Accuracy\\w.r.t. FCI (\%)} } & \textbf{Our Num Dets} & \textbf{Iterations} \\
					\hline
					H$_2$O  & cc-pVDZ & -76.232000  & -76.243900  & -76.243100  & -76.243405  & 99.998  & 1304036  & 8 \\
					H$_2$O  & 6-31G   & -76.113900  & -76.122370  & -76.120768  & -76.120905  & 99.997  & 213662   & 8 \\
					C$_2$   & cc-pVDZ & -75.666900  & -75.729840  & -75.726530  & -75.727818  & 99.995  & 180068  & 9 \\
					C$_2$   & 6-31G   & -75.584600  & -75.644180  & -75.642960  & -75.792674  & 99.998  & 93400   & 8 \\
					N$_2$   & cc-pVDZ & -109.243800 & -109.280700 & -109.268200 & -109.276482 & 99.988  & 1728927  & 9 \\
					N$_2$   & 6-31G   & -109.075900 & -109.108420 & -109.098980 & -109.104365 & 99.991  & 806720  & 9 \\
					\hline
				\end{tabular}
			}
			\caption{Comparison of Energy Approximations and Computational Effort: A Comparative Analysis between CISD, FCI, CCSD(T), and Our Proposed Model, Including the Number of Determinants and Iterations Required for Energy Convergence.}
			\label{energy_comparison}
		\end{table*}
		
		For instance, in the case of C$_2$ with the 6-31G basis, our model achieved 99.998\% accuracy with only 93,000 determinants, while CIPSI required 938,000 to reach 99.999\%. On average, our method reduced the number of determinants by 40.5\% compared to CIPSI, with savings ranging from 30\% to 50\%. 
		
		We also compared our method with the results reported by Basile and Herzog~\cite{basile_herzog}, who required 19 million determinants to reach 99.979\% of the correlation energy for N$_2$ with the cc-pVDZ basis. Our model achieved a comparable accuracy of 99.98\% using only 1.3 million determinants, representing a reduction by a factor of fifteen. Furthermore, under more conservative pruning settings, our approach reached the same 99.979\% target using just 137,412 determinants, highlighting the efficiency and scalability of our training strategy and pruning mechanism.

		We adjusted our model using a grid search, varying the number of epochs, temperature ($1/\beta$ in Eq. \ref{ecuacion_prob_to_hidden} and Eq. \ref{ecuacion_prob_to_visible}), batch size, and the number of Gibbs sampling steps used to approximate the hidden-visible distribution in the RBM. To illustrate the impact of temperature, Figure \ref{paralell_h2o} shows the results for the $H_2O$ molecule using the cc-pVDZ basis as a representative example. Overall, the results indicate that the model performs best at a temperature of $1/\beta=1$, suggesting that this value allows the RBM to effectively learn the underlying distribution. In contrast, higher temperatures ($1/\beta=5$ and $1/\beta=30$), which introduce greater stochasticity, led to consistently worse performance. In particular, the setting $1/\beta=30$, which corresponds to random sampling, resulted in the largest deviation from the FCI energy.\\
		
		Additionally, across all temperature settings, we observed that fewer epochs, smaller batch sizes, larger numbers of Gibbs sampling steps, and lower learning rates resulted in better performance compared to training for more epochs. This suggests that the model aims to approximate the overall distribution at each iteration, rather than overfitting to the current training set. Consequently, it maintains the ability to sample determinants beyond the training distribution, enhancing generalization. These findings are consistent with the patterns observed in Figure~\ref{paralell_h2o}, which summarizes the performance of various hyperparameter combinations.

		\begin{figure}[h]
			\begin{center}
				\includegraphics[width=.5\textwidth]{figure_4.png}
				\caption{Hyperparameter search results (only 4 iterations) showing energy (x axis) versus epochs (y axis) for $H_2O$ molecule in the cc-pVDZ basis for different temperatures $1/\beta=1,5$ and $30$. Models with lower temperatures and fewer epochs performed best, approaching FCI energy more closely.}
				\label{paralell_h2o}
			\end{center}
		\end{figure} 
		
\begin{figure}[h]
			\begin{center}
				\includegraphics[width=.47\textwidth]{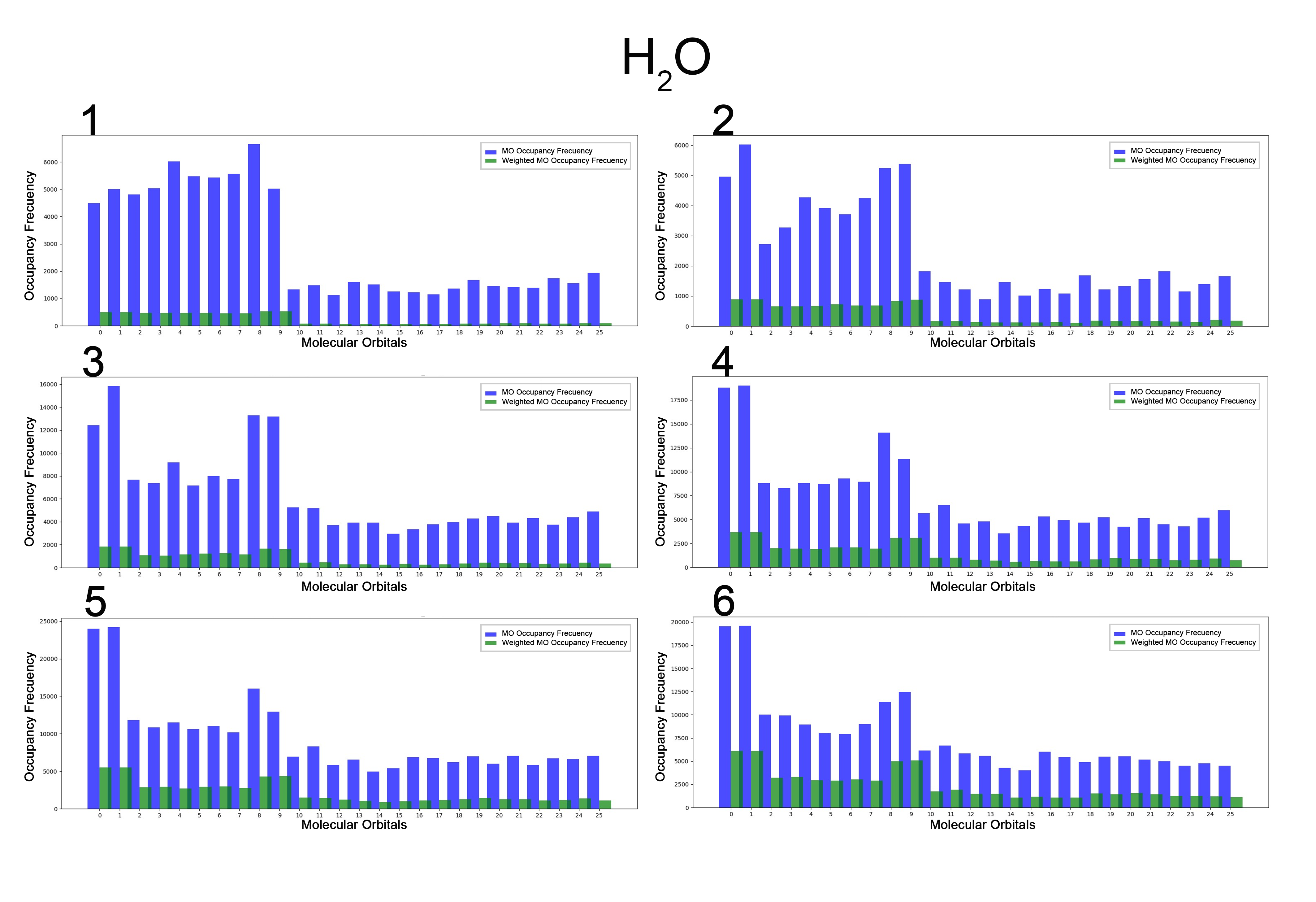}
				\caption{Estimated probability distribution of orbital occupation (blue bars), and the same distribution weighted by the coefficients obtained from diagonalization (green bars) for $H_2O$ in the 6-31G basis at each iteration.}
				\label{frec_ocup_h2o_631g}
			\end{center}
		\end{figure}

		Analyzing the estimated probability distribution of each Molecular Orbital (MO) for the water molecule in the 6-31G basis, Figure \ref{frec_ocup_h2o_631g} shows a transition from a Fermi-Dirac-like shape to a distribution predominantly concentrated in the lowest energy levels (first MOs for alpha and beta electrons), with a noticeable undulating pattern. This distribution is obtained by computing how frequently each MO is occupied during determinant generation at each iteration, and optionally weighting these frequencies by the squared coefficients ($|c_i|^2$) obtained from diagonalization. The resulting bar plots effectively capture the model’s learned sampling distribution for $H_2O$ in the 6-31G basis.
		
		These results may also explain the poor performance of high temperature models in Figure \ref{paralell_h2o}. RBMs with completely random training moves have difficulty exploring multimodal distributions efficiently \cite{Nicolas_2023}, leading to less effective sampling. This also clarifies why models trained with $1/\beta=1$ perform better with fewer epochs. Multimodal distributions are particularly challenging to sample because simulations can remain in metastable energy minima for extended periods, limiting the exploration of the configuration space. Training with fewer epochs helps mitigate this issue by reducing overfitting and improving coverage of relevant determinants.
		
		A similar trend appears when weighting the distributions by the coefficients obtained from diagonalization ($c_i$ in Eq.~\ref{equation_ci}), shown in green in Figure \ref{frec_ocup_h2o_631g}. These coefficients correspond to the squared amplitudes $|c_i|^2$ of each determinant and can be interpreted as their physical contribution to the total wavefunction, thus offering a probability-based view of orbital occupation.
		
\begin{figure}[h]
			\begin{center}
				\includegraphics[width=.47\textwidth]{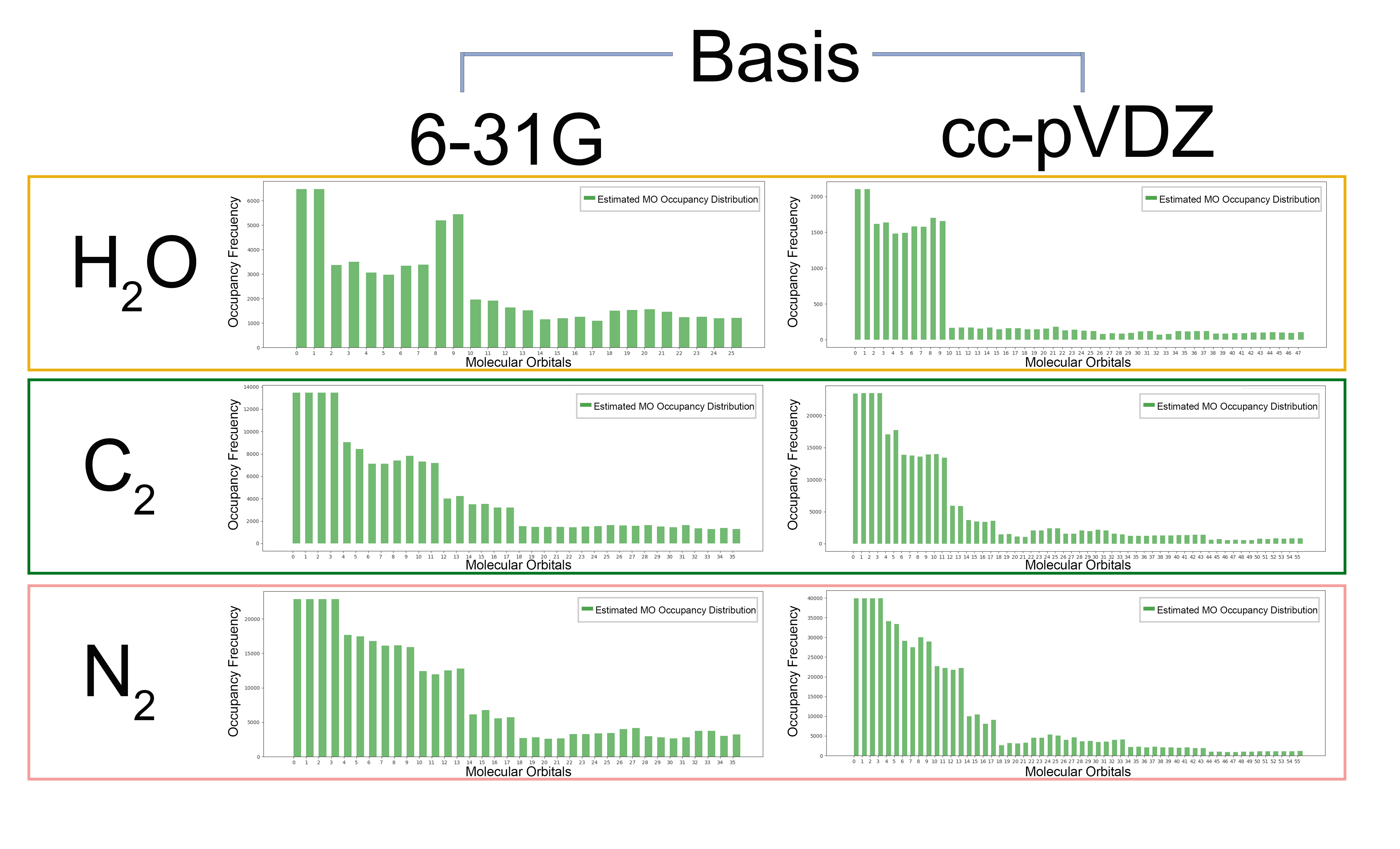}
				\caption{Final frequency occupation for each MO for for molecules (columns ) of  $H_2O$, $N_2$, and $C_2$ in cc-pVDZ and 631G basis (rows). }
				\label{all_frecuency}
			\end{center}
		\end{figure}

		Interestingly, the RBM consistently learns similar orbital occupancy patterns across multiple runs, even when starting from random initialization. While differences appear across different molecules, the same molecule across different basis sets exhibits recurring structural patterns. This is further illustrated in Figure \ref{all_frecuency}, where similar distributions emerge for $H_2O$ under different bases, while distinct molecules show more significant variations.
		
\begin{figure}[h]
			\begin{center}
				\includegraphics[width=.47\textwidth]{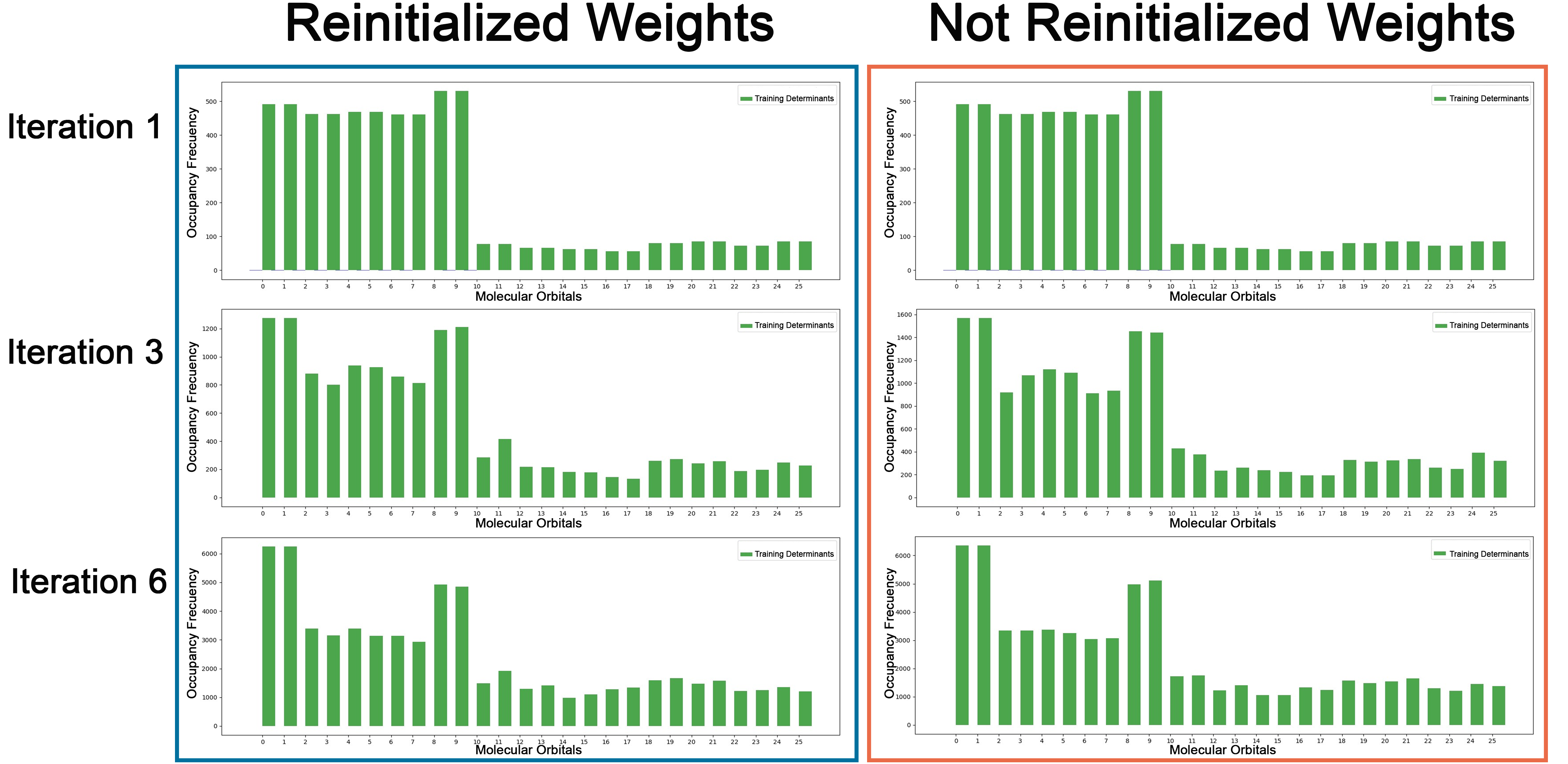}
				\caption{Comparison between frequency occupation in each Molecular Orbital reinitialized and not reinitialized in each iteration for $H_2O$ in 6-31G basis for each iteration.}
				\label{reinitialized_weights}
			\end{center}
		\end{figure}

		We verified that the RBM learns an underlying probability distribution by conducting two experiments on the same molecule. In one experiment, we randomly reinitialized the weights at each iteration, while in the other, the weights remained unchanged. Remarkably, both experiments produced distributions with the same behavior at each iteration (Figure \ref{reinitialized_weights}). This suggests that the RBM is not merely memorizing configurations, but is instead consistently identifying probability distribution of electrons in  Molecular Orbitals.
		
\begin{figure}[h]
			\begin{center}
				\includegraphics[width=.47\textwidth]{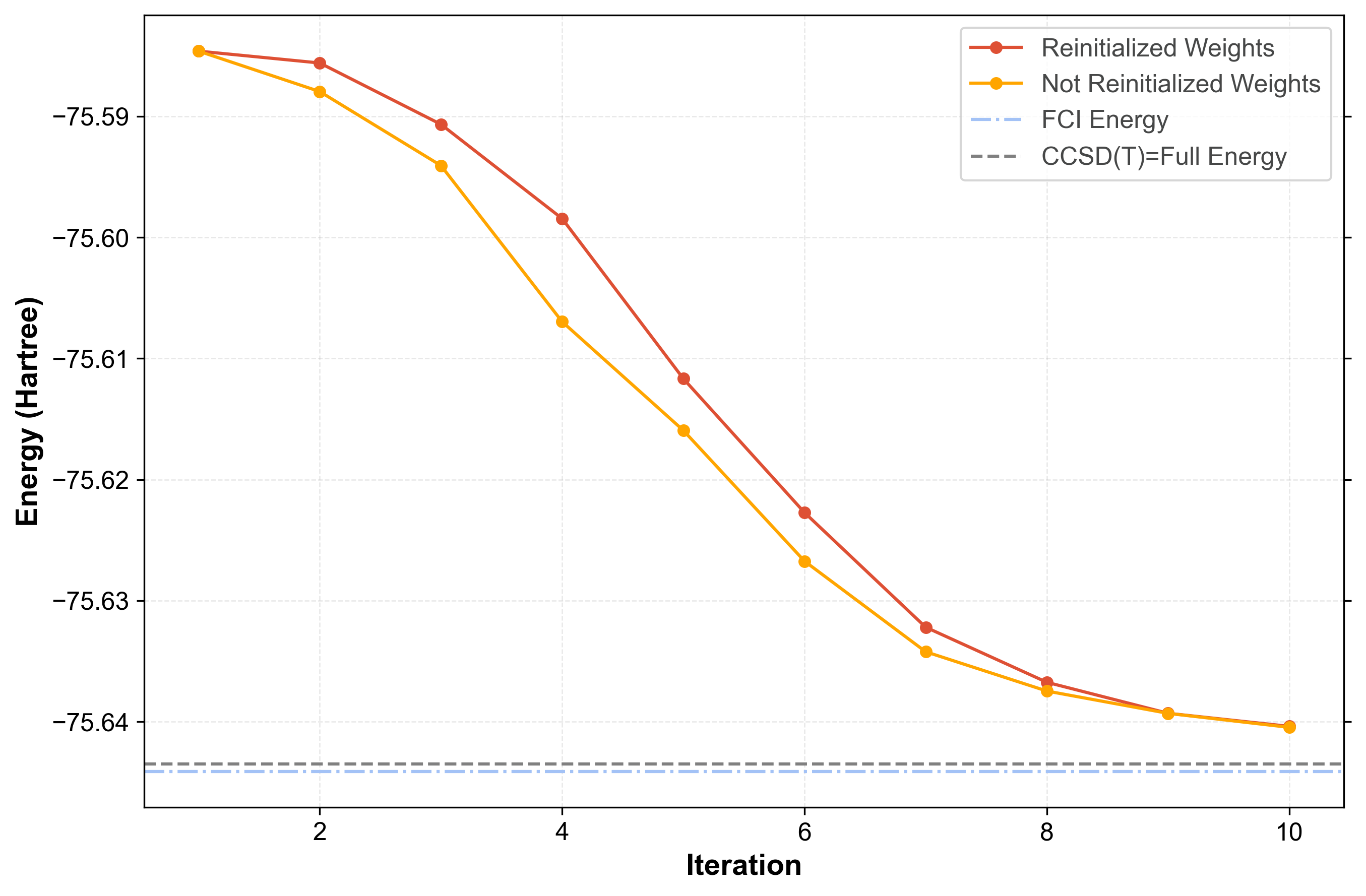}
				\caption{Energy convergence for the experiments with reinitialized and not reinitialized weight on each iteration for $C_2$ in 6-31G basis.}
				\label{e_converg_reinitialized_weights}
			\end{center}
		\end{figure}

		Figure \ref{e_converg_reinitialized_weights} shows that both experiments exhibit very similar energy convergence towards the FCI energy, but the experiment without reinitialization shows slightly faster convergence at early iterations. This suggests that avoiding reinitialization may provide an advantage in convergence speed. This strong similarity further supports the idea that the observed undulating pattern is not a feedback artifact generated by the RBM, but rather a learned feature of the system itself.

		The data from Figures \ref{all_frecuency} and \ref{reinitialized_weights} indicate that the model consistently converges to the same distribution, regardless of the molecular basis or weight initialization. This behavior suggests ergodic sampling, as the distribution continuously adjusts at each iteration.
		
		To formally verify that our sampling process is ergodic and does not produce non convergent Markov chains, we leverage the fact that the RBM is not trained via gradient descent, as in conventional neural networks, but instead through Gibbs sampling, a specific case of the Markov Chain Monte Carlo (MCMC) method \cite{JamesHeckman2001}. Given this formulation, we can apply the Gelman-Rubin statistic to assess convergence. Consider $j$ Markov chains, each initialized with different random values. For each chain $i$ of $L$ steps, the mean value of the chain is given by:
		
		\begin{equation}
			\bar{x}_j = \frac{1}{L} \sum_{i=1}^L x_{i}^{(j)}
		\end{equation}

		The mean of the means across all chains is:
		
		\begin{equation}
			\bar{\bar{x}}_j = \frac{1}{J} \sum_{j=1}^J \bar{x}_j
		\end{equation}
		
		The variance of the means of the chains, also known as the between chain variance \(B\), is defined as:
		
		\begin{equation}
			B = \frac{L}{J-1} \sum_{j=1}^J \left(\bar{x}_j - \bar{\bar{x}}_j\right)^2
		\end{equation}

		The average variance of the individual chains across all chains, known as the within chain variance \(W\), is:
		
		\begin{equation}
			W = \frac{1}{J} \sum_{j=1}^J \left( \frac{1}{L-1} \sum_{i=1}^L (x_{i}^{(j)} -\bar{x}_j )^2 \right)
		\end{equation}

		The Gelman-Rubin statistic \(R\) is then defined as \cite{gelman_rubin_estimator}:
		
		\begin{equation}
			R = \sqrt{\frac{\frac{L-1}{L}W + \frac{B}{L}}{W}}
		\end{equation}
		
		As \(L\) tends to infinity and \(B\) tends to zero, \(R\) tends to 1, indicating convergence.
		
		In our case, since the different runs of the algorithm converge to similar distributions after a certain number of iterations, and the energy values are consistent across these runs, it is as if we were running \(J\) simulations where the variance of the means of the chains is very low. For instance, the variance of the normalized distributions between the reinitialized and non-reinitialized weights is 3 $\times 10^{-4}$, which corresponds to the value of \(B\) in the Gelman-Rubin equation. This low variance implies that \(B\) is close to zero, leading to an estimate of the Gelman-Rubin statistic \(R\) that is very close to 1. This result suggests that our sampling process is ergodic and not subject to non convergence issues.

		\begin{figure*}[t]
			\begin{center}
				\includegraphics[width=0.9\textwidth]{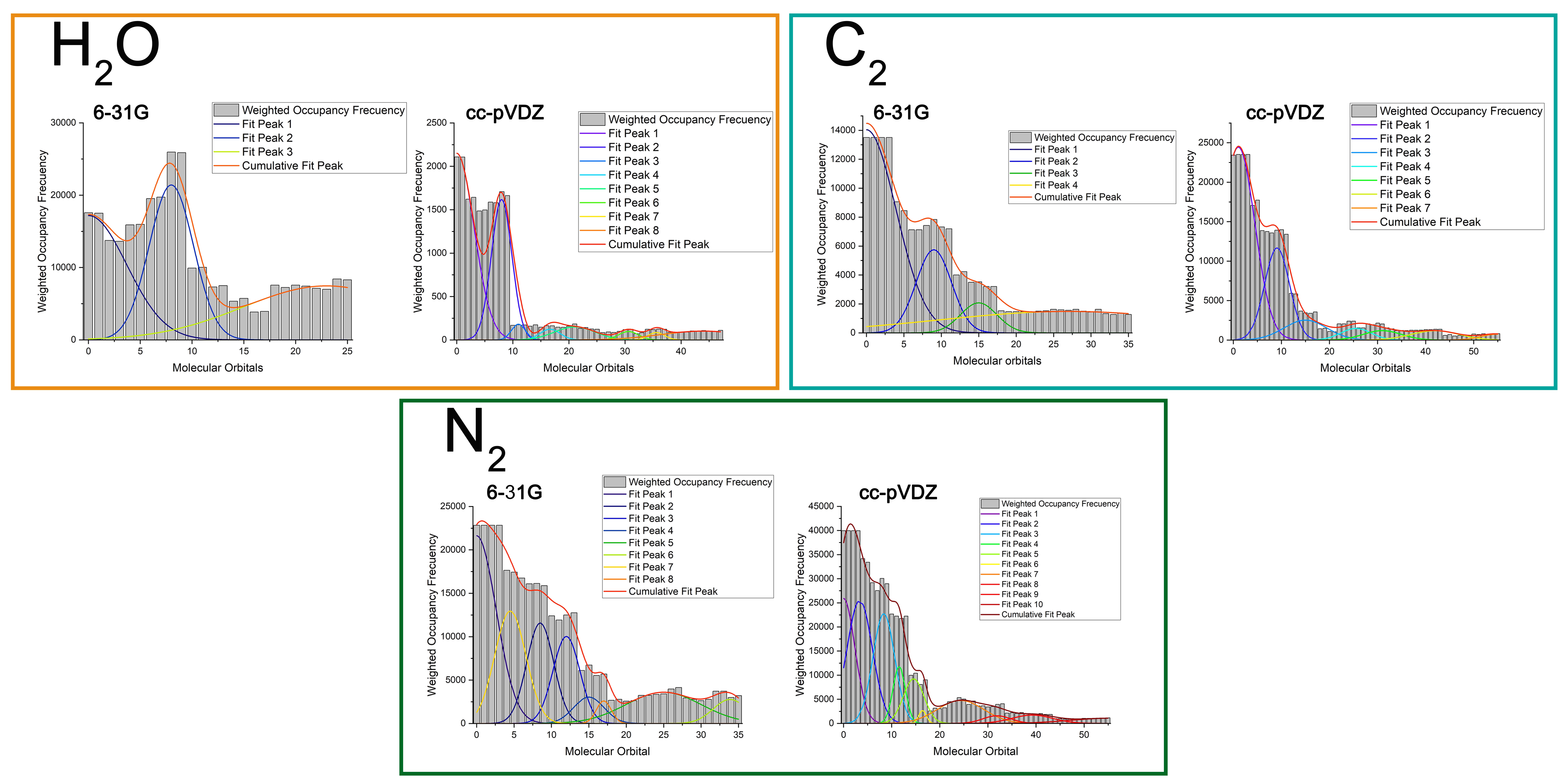}
				\caption{Gaussian distributions fitted to the weighted occupancy frequency graphs for different molecules and bases, illustrating similar behaviors. The Gaussian centers are conserved between bases for the same molecule, although the cc-pVDZ basis covers a larger MO space. Where they overlap, the patterns are comparable}
				\label{peaks_ajust_all}
			\end{center}
		\end{figure*}

		Another analysis confirming the RBM’s ability to learn consistent distributions is shown in Figure \ref{peaks_ajust_all}. Gaussian functions were fitted to the estimated probability distributions from Figure \ref{all_frecuency} to highlight general patterns. The resulting curves display similar behavior, with their means centered around the same molecular orbitals and showing consistent shapes across different basis sets. Naturally, the cc-pVDZ basis spans a larger number of orbitals than 6-31G, which explains the broader range observed in the distribution.

\begin{figure}[H]
			\begin{center}
				\includegraphics[width=.5\textwidth]{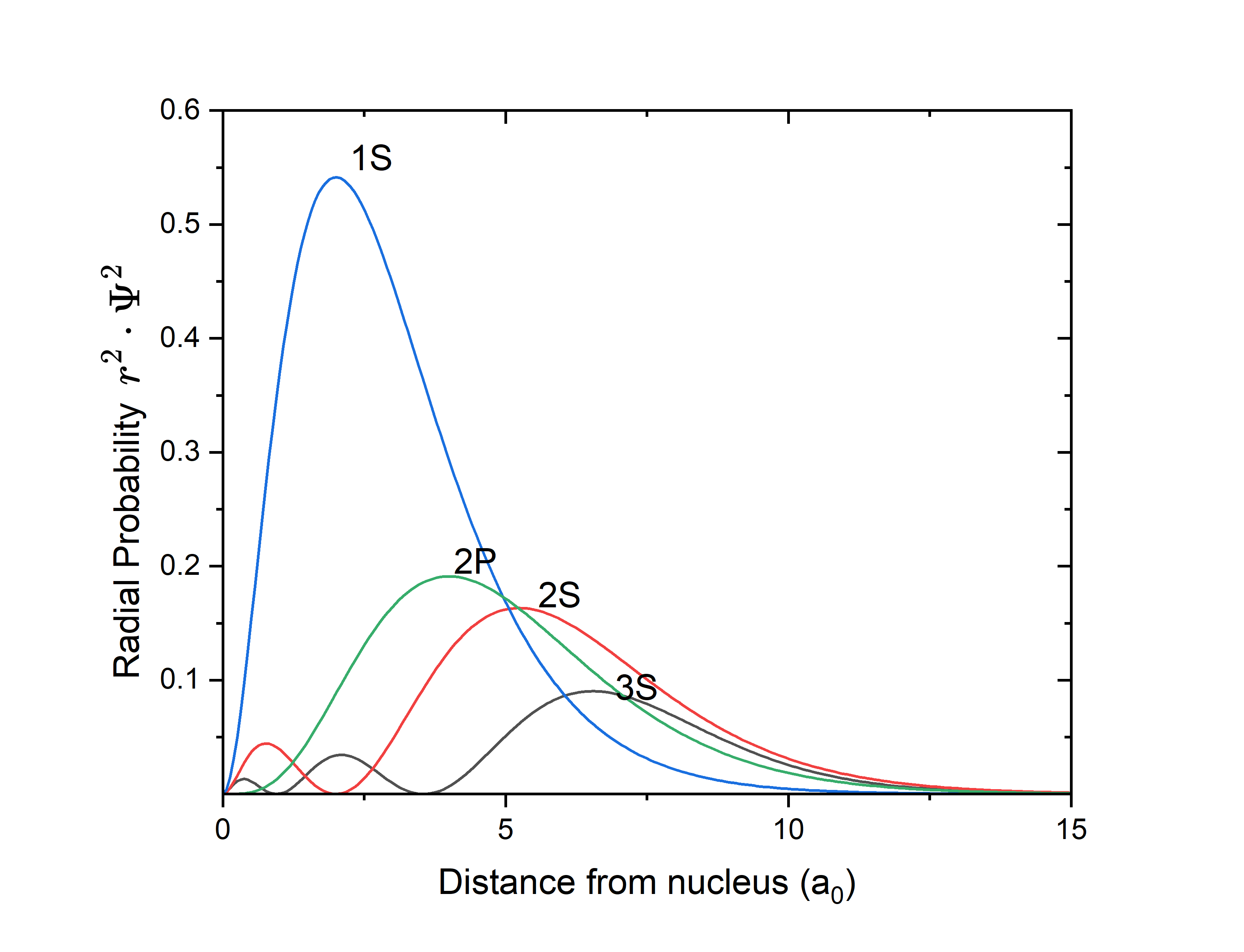}
				\caption{Radial Distribution Function (RDF) for a hydrogen atom, showing radial probability $\psi^2 r^2$ against distance $r$ from the nucleus in units of $a_0$. The graph displays the 1s, 2s, 2p, and 3s orbitals.}
				\label{rdf_generic}
			\end{center}
		\end{figure}

		Analyzing the behaviors of the envelope functions of Figure \ref{peaks_ajust_all}, a pattern emerges that resembles the radial probability distribution shown in Figure \ref{rdf_generic}. This plot shows a radial probability density $\psi^2 r^2$ against $r$, illustrating that electrons in higher energy orbitals are more likely to be found farther from the nucleus. The recurring modulations observed across different basis sets may suggest a connection to the molecule’s electronic configuration.
		
\begin{figure}[H]
			\begin{center}
				\includegraphics[width=.4\textwidth]{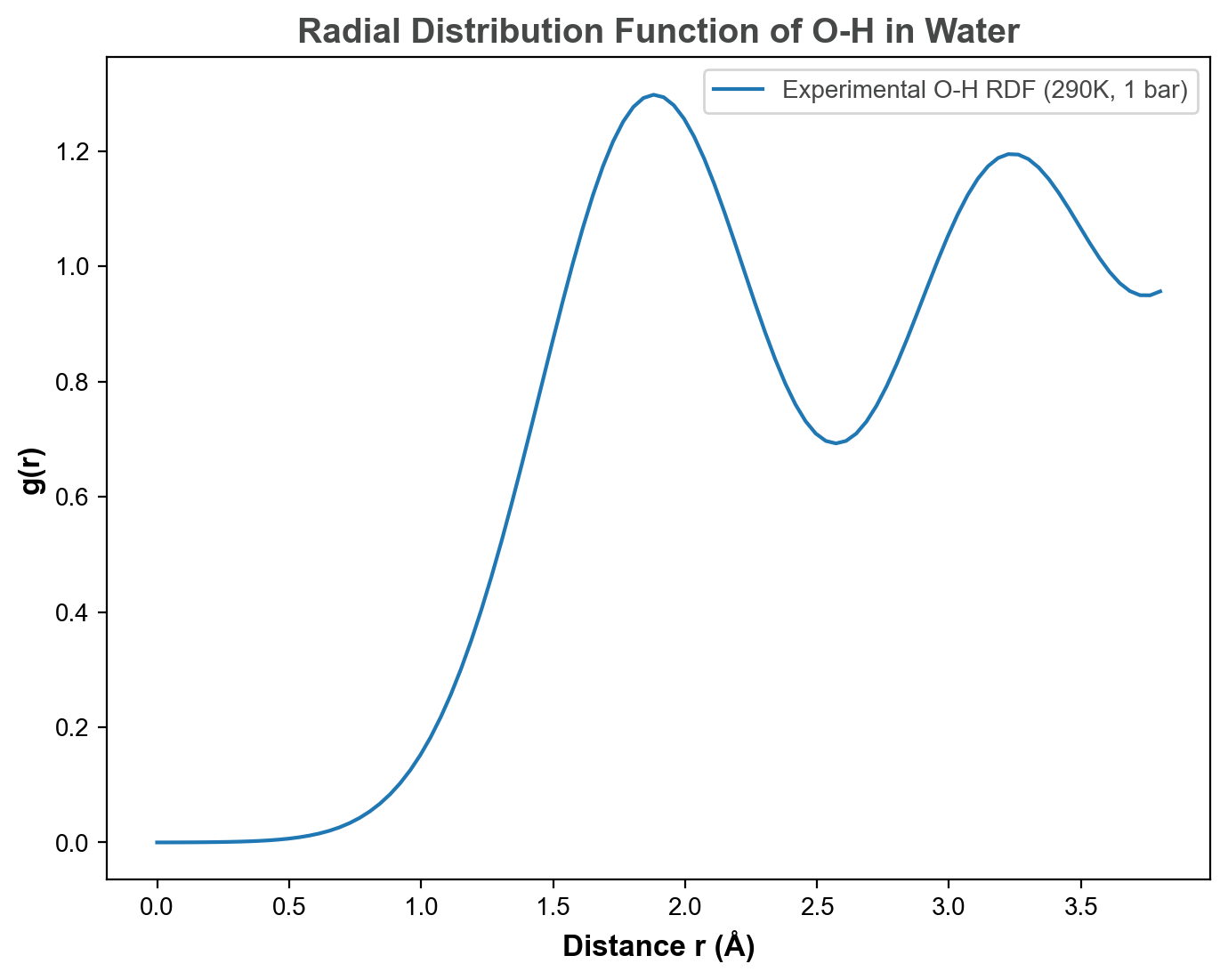}
				\caption{Radial Distribution Function (RDF) for water at 290K and 1 bar. The RDF was constructed using experimental and simulation data reported in \citep{h20_rdf_montecarlo_cordeiro, o_o_rdf_simulation_mcbride}.}
				\label{rdf_water}
			\end{center}
		\end{figure}

		We also observe a similar pattern to the RDFs obtained from experiments, rigid models, and Monte Carlo simulations for different molecules. Specifically, in Figure \ref{peaks_ajust_all}, the learned distribution for water molecule ($H_{2}O$) exhibit two prominent peaks that closely match those observed in the RDF of water at 290K and 1 bar (Figure \ref{rdf_water}), generated using experimental and simulation data \citep{o_o_rdf_simulation_mcbride, h20_rdf_montecarlo_cordeiro}. These peaks have been previously associated with the O-H bond structure. Similarly, for $C_2$, our model captures modulated patterns that align with RDF results reported in \citep{carbon_carbon_rdf_thompson, rdf_c2_koperwas}, while for $N_2$, the learned distributions show similarities with the RDF obtained under experimental conditions \citep{n_n_rdf_electric_field_cassone}. These observations reinforce the idea that the RBM effectively learns electron distribution patterns that reflect molecular bonding characteristics.

		While a complete correlation between our results and experimental RDFs cannot be established due to the inherent differences between atomic and molecular electron distributions, the observed similarities are notable. In atoms, electrons are distributed around a single nucleus in a predictable manner, whereas in molecules, neighboring atoms influence this distribution. These influences are three-dimensional and depend on the specific atomic positions. Although rules exist for combining atomic orbitals leading to constructive or destructive interference, these interactions manifest in complex three-dimensional spaces. Therefore, the applicability of a radial density function (RDF) may be limited, especially when compared across different molecules.\\

\begin{figure}[h]
			\begin{center}
				\includegraphics[width=0.48\textwidth]{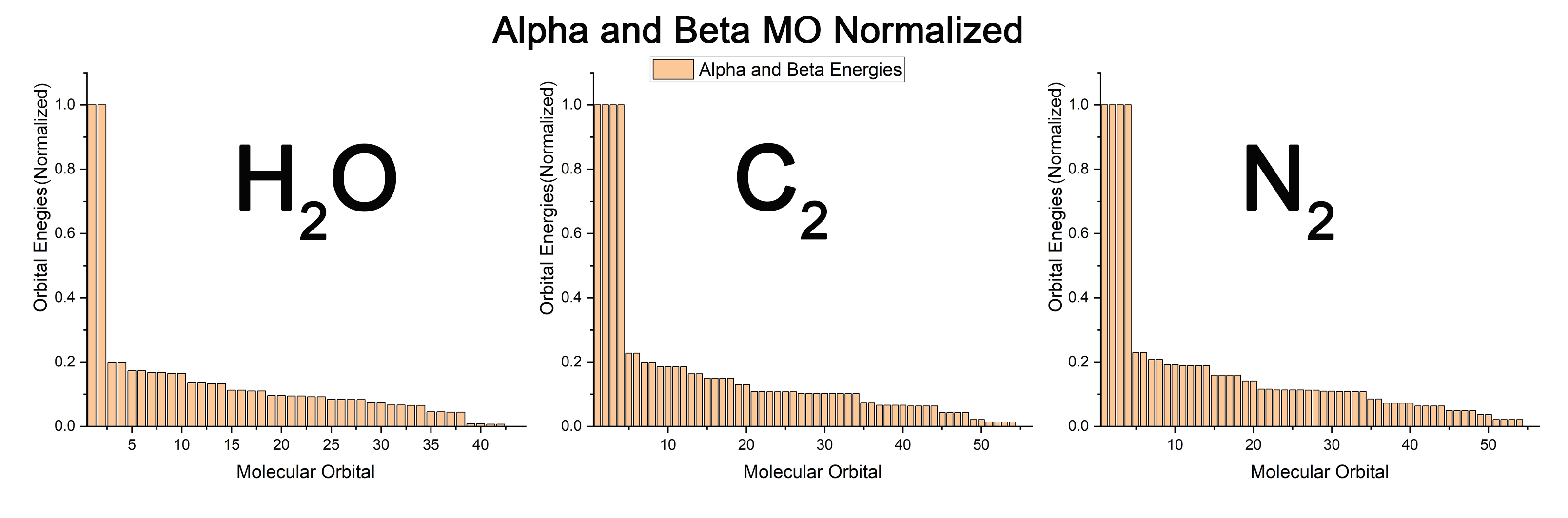}
				\caption{Comparison of the frequency distribution learned by the RBM with the normalized orbital energies for each molecule in the cc-pVDZ basis. The RBM captures distinct energy peaks, particularly in the lowest energy states, showing a remarkable correspondence with the computed orbital energies using Q-Chem \citep{qchem_software}.}
				
				\label{dist_vs_mo_energy}
			\end{center}
		\end{figure}

		Finally, an important analysis is observed in Figure \ref{dist_vs_mo_energy}, where we compare the distribution learned by the RBM with the normalized orbital energies for each molecule in the cc-pVDZ basis. The orbital energies were computed using the software Q-chem \citep{qchem_software}. Remarkably, the phenomenon of having two or four prominent energy peaks in the lowest states is observed in the distribution learned by the models (Fig \ref{peaks_ajust_all}). This suggests that the model might be learning relevant information about orbital energy, particularly for those with the lowest energy levels.
		
		\section{Conclusion}
		\label{Conclusions}
		We introduced an explainable and efficient method based on an RBM to approximate FCI calculations, significantly improving efficiency while preserving accuracy. Our approach represents an improvement over Quantum Monte Carlo (QMC) methods, offering a more structured and efficient way to explore determinant space. By effectively identifying and sampling the most relevant determinants, our model reduces the number of required determinants by up to four orders of magnitude compared to FCI and by 30–50\% compared to CIPSI, without compromising precision. The obtained energy values consistently align with those from highly accurate quantum chemistry methods, such as CCSD(T), demonstrating the robustness of our approach.

		Beyond its efficiency, our method provides a deeper understanding of the electronic structure. The RBM learns an intrinsic probability distribution over molecular determinants, capturing relevant quantum patterns. By analyzing the frequency of selected determinants, we observe a strong relationship between the learned distributions and molecular orbital occupancies, suggesting that the model is learning meaningful electronic structure features. Moreover, our results reveal undulating patterns in the learned determinant distributions that closely resemble Radial Distribution Functions (RDFs) associated with molecular bonding, further reinforcing the physical interpretability of our approach.
		
		Additionally, the ergodic nature of our sampling process was validated through statistical analysis, confirming that our RBM-based method does not produce non convergent Markov chains. This ensures that the learned distributions remain consistent across multiple runs, further supporting the robustness of our methodology.
		
		Our findings indicate that machine learning can play a crucial role in accelerating electronic structure calculations while preserving interpretability, offering a bridge between AI driven approaches and traditional quantum chemistry methods. Future work will explore extending this methodology to larger molecular systems and incorporating hybrid quantum and classical approaches to further improve efficiency and scalability.

		\section*{CRediT authorship contribution statement}
		\textbf{Jorge I. Hernandez-Martinez:} Conceptualization, Methodology, Software, Formal analysis, Investigation, Writing - Original Draft.
		
		\textbf{Gerardo Rodriguez-Hernandez:} Resources, Supervision, Writing - Review and Editing, Guidance in the physics domain.
		
		\textbf{Andres Mendez-Vazquez:} Resources, Supervision, Writing - Review and Editing, Guidance in neural networks.
		
		\textbf{Sandra Leticia Juárez-Osorio:} Review and Editing, Assistance in Software Development, and Support in Figure Preparation.

		\section*{Acknowledgements}
		
		This work was partially supported by CONAHCYT, Mexico

		\section*{Declaration of competing interest}
		The authors declare that they have no known competing financial interests or personal relationships that could have appeared to influence the work reported in this paper.

		\section*{Supplementary Information}
		
		A modular and fully reproducible implementation of the proposed method is available in \texttt{Python}. The source code and supplementary materials can be accessed at: \\
		\url{https://github.com/Navihdz/FCI-RBM}

	\appendix 
	\appendix
	\section*{Appendix A: Detailed Explanation of Restricted Boltzmann Machines}
	\label{Restricted_Boltzman_machine}
\begin{figure}[h]
		\begin{center}
			\includegraphics[width=.12\textwidth]{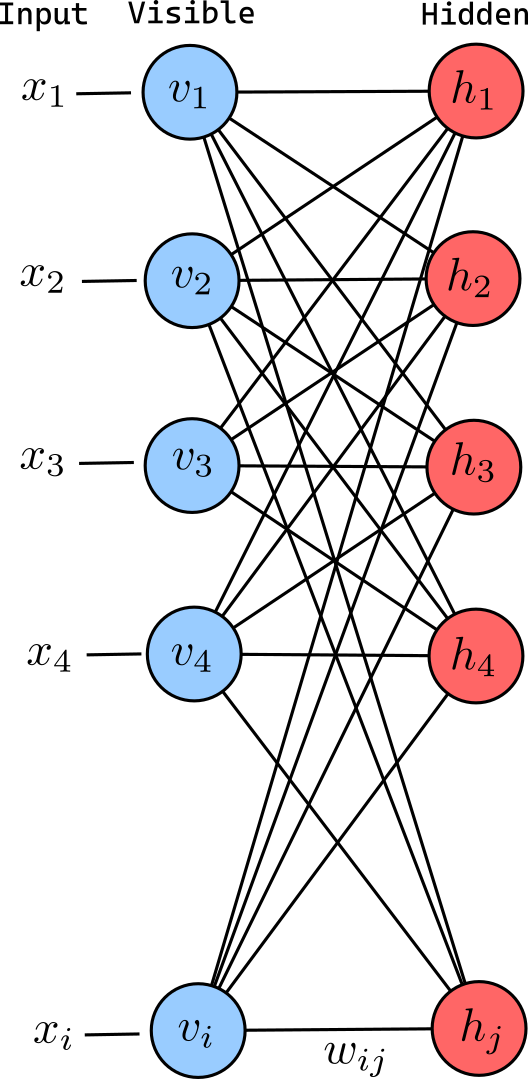}
			\caption{General architecture of the Restricted Boltzmann Machines.}
			\label{rbm_graph}
		\end{center}
	\end{figure}

	Restricted Boltzmann Machines is a type of generative model and unsupervised learning algorithm used in the field of machine learning.
	RBM is a probabilistic energy-based neural network consisting of two layers of nodes: a visible layer and a hidden layer. The nodes in the visible layer represent observed variables, while the nodes in the hidden layer represent latent or unobserved features. These two layers are fully connected to each other, but there are no connections within each layer (See Figure \ref{rbm_graph}).

	The main goal of an RBM is to learn the synaptic weights that maximize the joint probability of the observed data in the visible layer and the latent representations in the hidden layer, or in other words the RBM as a generative model, represents a
	probability distribution (over states of the visible units) where low energy configurations have higher probability. The energy is determined by the connection weights which are the parameters to be learnt from the data to built a generative model that learns the underlaying dynamic of the system \citep{articulo_overview_rbms}. This is achieved using an algorithm called ``Gibbs sampling" or ``Contrastive Divergence" \citep{article_hinton_training_rbms}.
	\\
	The learning process of an RBM involves initializing the synaptic weights randomly and propagating the data forward and backward in the network to calculate activations and reconstructions. The learning algorithm adjusts the synaptic weights to maximize the probability that the observed data is effectively reconstructed in the visible layer.
	\\
	
	A RBM can be mathematically described using the energy of a network and the associated probability distribution. The energy of an RBM is defined in terms of the activation states of the visible and hidden nodes, as well as the synaptic weights connecting these layers. The energy is denoted as $E(v,h)$, where $v$ represents the vector of states of the visible nodes and $h$ represents the vector of states of the hidden nodes.
	
	The energy of an RBM is defined as follows:
	
	\[
	E(v, h) = -\sum_{i} a_i v_i - \sum_{j} b_j h_j - \sum_{i,j} v_i h_j w_{ij}.
	\]
	
	Where:
	\begin{itemize}
		\item $a_i$ and $b_j$ are the biases of the visible and hidden nodes, respectively.
		\item $w_{ij}$ is the synaptic weight between visible node $i$ and hidden node $j$.
		\item $\sum_{i} a_i v_i$ and $\sum_{j} b_j h_j$ represent the contributions of the biases.
		\item $\sum_{i,j} v_i h_j w_{ij}$ represents the interactions between the visible and hidden nodes.
	\end{itemize}

	The network assigns a probability to every possible pair of a visible and a hidden vector via the energy function shown in Eq. \ref{equation_energy_function_rbm}:
	
	\begin{equation}
		p(v,h)=\frac{1}{Z} e^{-E(v, h)}.
		\label{equation_energy_function_rbm}
	\end{equation}
	This joint probability \(p(v, h)\) indicates the probability that both \(v\) and \(h\) are in a specific state.
	\\
	The probability distribution associated with an RBM is defined using the partition function $Z$, which normalizes the energy over all possible visible and hidden states:
	
	\[
	Z = \sum_{v} \sum_{h} e^{-E(v, h)},
	\]
	
	where the summations are performed over all possible visible and hidden states.
	\\
	
	Therefore, the probability assigned by the network to a visible vector $v$ is determined by summing over all possible hidden vectors $h$:
	
	\begin{equation}
		P(v) = \frac{1}{Z} \sum_{h} \exp(-E(v, h)).
	\end{equation}
	
	By adjusting the weights and biases, the network can increase the probability assigned to a training input. This adjustment lowers the energy of the target input while raising the energy of other inputs, particularly those with lower energies that significantly contribute to the partition function.
	
	Therefore, the parameters that maximize the following expression need to be found:
	\begin{equation}
		\underset{w,a,b}{\mathrm{argmax}}, \prod_{v \in V} p(v).
	\end{equation}
	
	To achieve this, it is necessary to maximize the expected value of $p(v)$:
	\begin{equation}
		\underset{w,a,b}{\mathrm{argmax}}, \mathbb{E}[\log p(v)].
	\end{equation}
	
	To do this, we take derivatives with respect to the weights and biases:
	
	\begin{equation}
		\frac{\partial}{\partial W} \log p(v),
		\label{derivada_log_w}
	\end{equation}
	\begin{equation}
		\frac{\partial}{\partial a} \log p(v),
	\end{equation}
	\begin{equation}
		\frac{\partial}{\partial b} \log p(v),
	\end{equation}
	
	The calculation of the derivative in Eq. \ref{derivada_log_w} can be computed as:

	\begin{equation}
		\begin{split}
			\frac{\partial \log p(v)}{\partial W} = &  \left[  \frac{\partial}{\partial W} (-E(v,h)) \vert v=v_n  \right] - \\
			&  \left[  \frac{\partial}{\partial W} (-E(v,h)) \right]\\
			& =\langle v_ih_j\rangle_{data}-\langle v_ih_j \rangle _{model}.
		\end{split}
	\end{equation}
	
	In this equation, $\mathbb{E}$ represents the expectation, and $\langle \cdot \rangle_{data}$ and $\langle \cdot \rangle_{model}$ denote expectations calculated under the data and model distributions, respectively.
	
	This results in a straightforward learning rule for performing stochastic steepest ascent in the log probability of the training data, as given by Eq. \ref{ecuacion_steepest_ascent}:
	
	\begin{equation}
		\Delta w_{ij}=\epsilon (\langle v_ih_j\rangle_{data}-\langle v_ih_j \rangle _{model}).
		\label{ecuacion_steepest_ascent}
	\end{equation}
	
	Because there are no direct connections between hidden units in an RBM, it is not necessary to know all the system configurations but rather the probabilities that connect a visible state to each of the nodes in the hidden state. Therefore, obtaining a sample of $\langle v_ih_j\rangle_{data}$ is very easy. Given a randomly selected training input, $v$, the binary state, $h_j$, of each hidden unit, $j$, is set to 1 with the following probability \citep{Hinton2012}:
	
	\begin{equation}
		p(h_j=1 \vert v)=\sigma (\beta(b_j+ \sum_i v_iw_{ij})).
		\label{ecuacion_prob_to_hidden}
	\end{equation}
	
	where $\sigma(x)$ is the logistic function and $\beta$ is the inverse of temperature ($1/t$) and is used to control the noise in the sampling, this means that when $t\rightarrow \infty$ all the visible probabilities  given $h_j=0$ or $1$ will be equal to one half.
	
	Similarly, as there are no connections between the visible nodes, we can obtain a sample of $\langle v_ih_j \rangle_{data}$ given a hidden vector by setting each visible unit $v_i$ to 1 with the probability given by Eq. \ref{ecuacion_prob_to_visible}:
	
	\begin{equation}
		p(v_i=1 \vert h)=\sigma (\beta (a_i+ \sum_j h_jw_{ij})).
		\label{ecuacion_prob_to_visible}
	\end{equation}
	
	The problem arises when trying to obtain the expectation under the model distribution $\langle v_ih_j \rangle_{model}$ since calculating the expected value over the entire model would require knowing all possible configurations between visible and hidden states in order to obtain the probability of each one and compute the expected value \cite{Hinton2012}. This problem quickly becomes intractable \citep{articulo_overview_rbms} . For example, in an RBM with 100 nodes in the visible layer and 200 nodes in a hidden layer, there would be $2^{300}$ possible configurations, a number greater than the number of atoms in the universe ($10^{78}-10^{82}$ atoms in the universe). Storing all that information is impossible, even for a relatively simple network like the one described above.
	
	This problem can be solved by starting at any random state of the visible units and performing alternating Gibbs sampling for a very long time. Gibbs sampling is a Markov chain Monte Carlo algorithm where each iteration consists of updating all the hidden units in parallel using Eq. \ref{ecuacion_prob_to_hidden}, followed by updating all the visible units in parallel using Eq. \ref{ecuacion_prob_to_visible}.
	
	A much faster learning procedure called Contrastive Divergence was proposed by Hinton \citep{article_hinton_training_rbms}. It starts by setting the states of the visible units to a training vector. Then, the binary states of the hidden units are all computed in parallel using Eq. \ref{ecuacion_prob_to_hidden}. Once the binary states have been determined for the hidden units, a  ``reconstruction" is produced by setting each $v_i$ to 1 with a probability given by Eq. \ref{ecuacion_prob_to_visible}. The change in a weight is then given by:
	
	\begin{equation}
		\Delta w_{ij}=\epsilon (\langle v_ih_j\rangle_{data}-\langle v_ih_j \rangle _{recon}).
		\label{delta_weight}
	\end{equation}
	
	In this equation, $\epsilon$ represents a learning rate, and $\langle \cdot \rangle_{recon}$ denotes expectations calculated using the reconstruction data.
	
	RBMs typically learn better models if more steps of alternating Gibbs sampling are used before collecting the statistics for the second term in the learning rule in Eq. \ref{delta_weight} , which will be called the negative statistics \citep{Hinton2012}.
	
	Contrastive Divergence is an algorithm to calculate maximum likelihood in order to estimate the slope of the graphical representation showing the relationship between the weights and its errors in a time complexity of each step of $\mathcal{O}(1)$.
	
	Without Contrastive Divergence the hidden nodes in the restricted Boltzmann machine can never learn to activate properly, this method is also used in the cases where the direct probability can’t be evaluated, also, this method is one of the fastest methods to measure the log partition function.

	While Contrastive Divergence facilitates efficient learning in RBMs, their true power lies in their representational capacity, enabling them to capture a wide range of distributions. Increasing the number of hidden units enhances performance by improving training log likelihood and reducing the KL divergence between data and model distributions \citep{representational_power_rbms}. Research has shown that an RBM with $k+$1 hidden units can closely approximate any distribution over $\{0,1\}^n$, where $k$ denotes the number of input vectors with non-zero probability. Furthermore, it has been demonstrated that an RBM with $2^n-1$ hidden units can approximate any distribution \citep{article_refinements_rbms}. Refinements in these findings suggest that RBMs with $\alpha2^n$-1 hidden units (where $\alpha < 1$) are sufficient for approximating any distribution \citep{article_hierarchical_models}. This versatility makes RBMs particularly promising for addressing challenges such as FCI in quantum mechanics, where the curse of dimensionality severely limits computational tractability for complex systems. By leveraging their ability to capture complex distributions efficiently, RBMs offer a potential solution to overcoming the computational challenges posed by FCI.

	Here's how RBMs can bridge this gap:

	\begin{itemize}
		\item \textbf{Learning the Configuration Space:} An RBM can be trained on a dataset of electronic configurations relevant to the system under study. By learning the underlying probability distribution of these configurations, the RBM can efficiently sample the most important configurations for the FCI calculation.
		
		\item \textbf{Selective Sampling vs. Random Exploration:} Traditional montecarlo sampling based methods for FCI might explore the configuration space randomly, potentially wasting computational resources on irrelevant configurations. The RBM, on the other hand, prioritizes generating new configurations, giving higher likelyhoood to those that contribute more to the solution. This selective approach significantly reduces the computational cost of FCI.
		
	\end{itemize}
	

	\bibliographystyle{plainnat}

	\bibliography{ref}

\end{document}